\documentclass[conference]{IEEEtran}
\IEEEoverridecommandlockouts
\usepackage{cite,multirow,balance}
\usepackage{graphicx,listings,longtable}
\usepackage{subfig} 

\usepackage{amsmath,amssymb,amsfonts}
\usepackage{algorithmic}
\usepackage{graphicx}
\usepackage{textcomp}
\usepackage{xcolor}
\def\BibTeX{{\rm B\kern-.05em{\sc i\kern-.025em b}\kern-.08em
    T\kern-.1667em\lower.7ex\hbox{E}\kern-.125emX}}

\begin{document}





\title{Conceptual Metaphor Theory as a Prompting Paradigm for Large Language Models}

\author{
\IEEEauthorblockN{Oliver Kramer}
\IEEEauthorblockA{
Computational Intelligence Group \\
Department of Computer Science \\
Universit\"at Oldenburg, Germany \\
Email: oliver.kramer@uni-oldenburg.de
}}

\maketitle

\begin{abstract}
We introduce Conceptual Metaphor Theory (CMT) as a framework for enhancing large language models (LLMs) through cognitive prompting in complex reasoning tasks. CMT leverages metaphorical mappings to structure abstract reasoning, improving models' ability to process and explain intricate concepts. By incorporating CMT-based prompts, we guide LLMs toward more structured and human-like reasoning patterns.
To evaluate this approach, we compare four native models ({Llama3.2}, {Phi3}, {Gemma2}, and {Mistral}) against their CMT-augmented counterparts on benchmark tasks spanning domain-specific reasoning, creative insight, and metaphor interpretation. Responses were automatically evaluated using the {Llama3.3} 70B model. Experimental results indicate that CMT prompting significantly enhances reasoning accuracy, clarity, and metaphorical coherence, outperforming baseline models across all evaluated tasks.
\end{abstract}

\begin{IEEEkeywords}
Conceptual Metaphor Theory, Large Language Models, Prompt Engineering, Benchmark Evaluation, Complex Reasoning
\end{IEEEkeywords}

\section{Introduction}

Large Language Models (LLMs) excel in various domains, including natural language understanding and complex problem-solving. However, enhancing their reasoning capabilities across diverse tasks remains challenging. Integrating insights from Conceptual Metaphor Theory (CMT), which emphasizes understanding abstract concepts through metaphorical mappings from concrete domains, offers a promising approach. By designing prompts to incorporate metaphoric reasoning, LLMs can be guided to tackle complex reasoning tasks more effectively.

In this study, we propose a CMT-inspired prompt engineering strategy to enhance the reasoning capabilities of LLMs. We hypothesize that LLMs can generate more coherent, contextually rich, and insightful responses across diverse task categories using CMT-inspired prompts.
We evaluate this approach on a benchmark dataset containing 100 tasks from multiple reasoning domains, including domain-specific reasoning, creative insight puzzles, and general problem-solving. Using four advanced LLMs -- {Llama3.2}, {Phi3}, {Gemma2}, and  {Mistral} -- we rigorously assess the effectiveness of CMT-inspired prompts compared to baseline prompts in terms of accuracy, coherence, and depth of reasoning.

The findings from this study highlight the potential of metaphor-guided prompts as a transformative tool for enhancing reasoning in LLMs, providing actionable insights for the development of more capable and contextually aware AI systems.

The paper is structured as follows: Section~\ref{sec:cmt} introduces CMT, followed by Section~\ref{sec:cmt_prompting}, which details the CMT-based prompting approach. Section~\ref{sec:related} reviews related work, while Section~\ref{sec:bench} presents the benchmark problems. Section~\ref{sec:exp} describes the experimental setup and results. Finally, Section~\ref{sec:cons} discusses key findings and outlines future research directions.

\section{Conceptual Metaphor Theory}
\label{sec:cmt}

CMT is a foundational concept in cognitive science that explores how humans understand and process abstract concepts through metaphoric mappings from more tangible, familiar domains. Introduced by George Lakoff and Mark Johnson \cite{Lakoff1,Lakoff2}, CMT posits that metaphors are not merely linguistic devices but cognitive tools deeply embedded in human thought and reasoning. They enable individuals to comprehend complex or abstract ideas by linking them to more accessible and experiential domains. For example, the metaphor “time is money” frames the abstract concept of time using the concrete and familiar framework of financial transactions.

Central to CMT is the concept of \textit{source domains} and \textit{target domains}, which form the foundation of metaphorical thinking. The source domain represents a familiar, concrete concept grounded in physical experience or everyday life, such as spatial orientation, movement, or physical objects. In contrast, the target domain encapsulates an abstract, intangible idea that is often more challenging to conceptualize directly, such as emotions, time, morality, or complex systems. Through systematic mappings between these domains, specific properties, relationships, or structures from the source domain are transferred to the target domain. These mappings enable individuals to use their intuitive understanding of the source domain to make sense of abstract phenomena in the target domain.

For example, in the metaphor “time is a resource,” the source domain of resources (e.g., money or energy) lends its structure and relationships—such as scarcity, management, or expenditure—to the target domain of time. This allows people to conceptualize time as something that can be “spent,” “saved,” or “wasted,” making an abstract concept more tangible and actionable. Similarly, in the metaphor “knowledge is a journey,” the spatial and physical aspects of a journey—such as paths, destinations, obstacles, and progress—are mapped onto the process of acquiring knowledge, enabling individuals to conceptualize learning as a dynamic, goal-oriented activity.

These mappings are not arbitrary; they are shaped by embodied experiences and cultural context, which influence how individuals and societies construct and interpret metaphors. The systematic transfer of relationships and properties from source to target domains is what makes metaphors powerful tools for learning, problem-solving, and creativity. By framing unfamiliar or abstract concepts in terms of familiar ones, metaphors provide cognitive shortcuts that reduce complexity, aid memory, and enhance communication.

This central mechanism of CMT highlights the profound role of metaphor in shaping human thought, reasoning, and language. It also serves as the basis for applying CMT to computational systems, where metaphoric mappings can be leveraged to enhance the reasoning capabilities of artificial intelligence, particularly in tasks requiring abstraction and analogical thinking.

CMT has been extensively studied across disciplines, including linguistics, psychology, and education, as a tool for enhancing human learning and problem-solving. Its influence extends to computational applications, where metaphors have been used to structure knowledge representation and reasoning in artificial systems. However, its potential remains underexplored in the realm of large language models, despite their impressive capabilities in generating human-like text.

\section{CMT Prompting Approach}
\label{sec:cmt_prompting}

The CMT prompting approach employs structured metaphorical reasoning through CoT prompting. This method enables LLMs to systematically interpret abstract concepts (target domains) by mapping them to familiar physical experiences (source domains). By structuring the reasoning process step by step, these models emulate human cognitive patterns, enhancing their ability to generate more insightful and contextually relevant responses. The approach is configured within Ollama models, with the configured versions referred to as {CMT-Llama3.2}, {CMT-Phi3}, {CMT-Gemma}, and {CMT-Mistral}.

\subsection{LLM Configuration}

The CMT prompting approach relies on three key components: source and target domains, inference through metaphorical mappings, and systematic reasoning. The system is guided by the instructions, see Figure \ref{fig:instructions}.

\begin{figure}[htb]
\begin{lstlisting}
# Set the temperature to 0.7 for balanced creativity and coherence
PARAMETER temperature 0.7

# Define the system message to guide the model's behavior
SYSTEM
As a cognitive agent utilizing Conceptual Metaphor Theory (CMT), you can interpret abstract concepts (target domains) through more concrete experiences (source domains). This approach facilitates understanding by mapping familiar physical experiences onto intangible ideas.

Source and Target Domains:
- Source Domain: The concrete or physical experience from which we draw metaphorical expressions.
- Target Domain: The abstract concept we aim to understand through the source domain.

Inference Process:
By mapping elements from the source domain onto the target domain, you can infer characteristics of the abstract concept based on the concrete experience. This mapping allows for a structured understanding of the target domain.
\end{lstlisting}
\caption{Instructions for configuration of CMT-prompted LLMs\label{fig:instructions}}
\end{figure}

\subsection{Inference Through Source-Target Mappings}

The CMT-prompted models leverage systematic source-target mappings to interpret and explain abstract concepts. This process involves identifying a relevant source domain, transferring its properties and relationships to the target domain, and constructing a coherent explanation or solution. These mappings, integrated as CMT-prompts in an LLM-configuration process, enable the models to internalize metaphorical reasoning and apply it dynamically across tasks.

This structured mapping process follows CoT-like approach within the CMT framework, guiding the model step-by-step through conceptual associations to derive logical inferences. By explicitly encoding the reasoning steps that bridge the source and target domains, CMT-prompted models develop a more structured interpretive process, much like CoT enables stepwise problem-solving in mathematical and logical tasks.

The following examples illustrate this structured mapping process, which is incorporated into the CMT prompt using a CoT-like approach, see Figure \ref{fig:cot}.

\begin{figure}[htb]
\begin{lstlisting}
	1.	Time is money:
	- Source Domain: Money (a valuable resource)
	- Target Domain: Time
	- Inference: Just as money is valuable and should be spent wisely, time is also valuable and should be used efficiently
    
	2.	He has a heart of stone:
	- Source Domain: Stone (hard and unfeeling)
	- Target Domain: A person's heart (emotions)
	- Inference: The person is unfeeling or unsympathetic, similar to the hardness of stone
    
	3.	The world is a stage:
	- Source Domain: Stage (platform for performances)
	- Target Domain: The world/life
	- Inference: Life is like a play where people have roles to perform, suggesting that daily activities are performances
\end{lstlisting}
\caption{CMT-inspred CoT\label{fig:cot}}
\end{figure}

By embedding these examples in the CMT-prompts, the models acquire the ability to generalize structured metaphorical reasoning across a wide range of abstract and domain-specific tasks, enhancing their interpretive and explanatory capabilities.

\subsection{Implementation Details}

The CMT-prompted models operate with a balanced temperature parameter ({temperature=0.7}) to ensure a mix of creativity and coherence. The models are preconditioned with a system message that emphasizes their role as cognitive agents employing metaphorical reasoning.

The inclusion of these principles within the configuration process eliminates the need for manually crafting explicit CMT-based prompts for every task. Instead, the models are inherently capable of performing structured metaphorical reasoning, making them versatile for a wide range of applications, including problem-solving, teaching, and interpretation.

\section{Related Work}
\label{sec:related}

\subsection{Prompt Engineering}

Prompt engineering has become a crucial technique for enhancing the reasoning capabilities of LLMs. Various strategies have been developed to improve model performance across different types of reasoning tasks. Zero-shot prompting generates responses without prior examples, relying on the model’s pre-trained knowledge, while few-shot prompting \cite{brown2020language} improves accuracy by incorporating task-specific examples within the prompt. More structured approaches, such as Chain-of-Thought (CoT) prompting \cite{wei2022chain}, further enhance reasoning by explicitly breaking down complex problems into sequential steps. Extending this idea, Tree of Thoughts (ToT) prompting \cite{tree} allows the model to explore multiple reasoning paths in parallel, improving decision-making in tasks that benefit from divergent thinking. Another related approach, ReAct \cite{yao2022react}, integrates logical reasoning with interactive decision-making, making it particularly useful for dynamic environments.

Beyond structured reasoning, several methods focus on optimizing prompts for improved performance. Prompt Breeder \cite{promptbreeder2022} applies evolutionary strategies to refine prompts iteratively, while Automated Prompt Engineering (APE) \cite{ape} and Optimization by PROmpting (OPRO) \cite{opro} automate the prompt design process through reinforcement learning and search-based techniques. These methods systematically fine-tune instructions to maximize the model's effectiveness, reducing reliance on manual prompt crafting. 

Cognitive Prompting \cite{cp} introduces structured cognitive operations to enhance problem-solving abilities in LLMs, incorporating step-by-step reasoning techniques to improve multi-step task performance. By guiding models through explicit reasoning structures, these approaches demonstrate notable improvements, particularly in arithmetic and logical reasoning benchmarks.
 
\subsection{LLMs and Conceptual Metaphor Theory}

Recent work has explored prompting techniques to improve LLMs’ understanding of metaphors. Prystawski et al. \cite{prystawski2023psychologically} employ CoT-style prompts to guide models in analyzing metaphorical language, showing improved interpretative accuracy. Comsa et al. \cite{Comsa} introduce MiQA, a benchmark assessing LLMs’ ability to reason with conventional metaphors, combining metaphor detection with commonsense reasoning. Hicke and Kristensen-McLachlan \cite{Hicke} evaluate models’ capabilities in systematically recognizing and annotating conceptual metaphors using linguistic annotation frameworks.

While these studies demonstrate progress in metaphor understanding, they primarily focus on recognition and classification rather than structured conceptual reasoning. This work builds on previous approaches by designing tasks that explicitly incorporate conceptual mappings, evaluating how well LLMs establish connections between source and target domains. By assessing models’ ability to apply metaphorical structures across diverse problem types, this study provides a more comprehensive evaluation of metaphor-driven reasoning in LLMs.

\section{Benchmark Problems}
\label{sec:bench}

To evaluate the ability of LLMs to comprehend, interpret, and reason using metaphorical language, we developed a benchmark comprising 100 problems. These problems are categorized into four distinct groups, each designed to assess a different facet of metaphor understanding, generation, and application. The benchmark examines both the recognition of existing metaphors and the capacity to construct novel conceptual mappings for abstract reasoning and explanation. Within each category, we created four specialized tasks to ensure a comprehensive evaluation.
To preserve the integrity of the benchmark, task instructions do not explicitly prompt the models to identify or rely on metaphorical reasoning.

\subsection{Metaphor Identification and Mapping (MIM)}

Metaphors play a fundamental role in human cognition, structuring abstract concepts through more concrete experiences. 

Each task presents a sentence or passage in which a conceptual metaphor is embedded. The model must determine the key entities involved, recognize the source and target domains, and explain how the metaphor frames the underlying meaning. For instance, given the statement \emph{“Their plan to revitalize the industry took root and began to grow stronger each year”}, an effective response should identify how \emph{growth} in a biological sense is mapped onto the success of an economic or organizational strategy.

Performance is evaluated across three dimensions: (1) \textit{Precision in structural interpretation}, assessing how well the response captures the conceptual elements of the metaphor; (2) \textit{Coherence of explanation}, measuring logical consistency and insightfulness; and (3) \textit{Accuracy in mapping relationships}, determining whether key relationships between entities are correctly identified. This structured evaluation allows for a fine-grained comparison between baseline LLMs and models enhanced with CMT-based reasoning strategies.

\subsection{Domain-Specific Reasoning (DSR)}

DSR tasks challenge the model to apply metaphorical reasoning in specialized fields, requiring precise conceptual mappings beyond conventional analogies. Unlike general metaphor identification tasks, these problems demand a structured explanatory approach, where the model must construct meaningful analogies that enhance comprehension of complex topics.

Each task presents an abstract or technical concept drawn from disciplines such as physics, biology, economics, artificial intelligence, and engineering. The model must provide an explanation that not only adheres to the principles of the domain but also introduces a novel, insightful metaphor that facilitates understanding. Traditional explanations, such as likening an electric circuit to water flowing through pipes, are insufficient. Instead, the model must generate alternative mappings that reveal deeper conceptual structures. For instance, explaining market competition should go beyond the common battlefield metaphor and explore alternative frameworks, such as ecological ecosystems or evolutionary dynamics.

The evaluation criteria for DSR tasks focus on three core aspects: (1) \textit{Precision in structural interpretation} – assessing how well the response captures the key conceptual elements; (2) \textit{Coherence of explanation} – evaluating the logical consistency and insightfulness of the account; and (3) \textit{Accuracy in mapping relationships} – measuring the correct establishment of key relationships between described entities. These criteria ensure that responses are not only metaphorically rich but also technically sound.

\subsection{Explanation and Teaching Tasks (ETT)}

ETT evaluate a model’s ability to communicate complex concepts to a general audience using clear and intuitive analogies. These tasks focus on how effectively an AI system can translate specialized knowledge into accessible explanations that facilitate understanding. Instead of merely providing factual responses, the model must construct meaningful metaphors that enhance comprehension for non-expert readers.

The problem set includes a diverse range of domains, spanning artificial intelligence, economics, physics, biology, and social sciences. For instance, an ETT task may ask the model to explain how neural networks learn using an analogy, such as comparing them to a child learning through trial and error. Similarly, an economic concept like inflation may be illustrated through a metaphor likening it to air being pumped into a balloon, causing it to expand over time. The model's explanations are assessed based on their clarity, conceptual fidelity, and the effectiveness of the chosen metaphor in conveying the underlying principles.

Each task is evaluated according to three core criteria: (1) \textit{Clarity for Non-Experts}, which measures how well the response avoids unnecessary complexity while maintaining accuracy; (2) \textit{Conceptual Accuracy}, assessing whether the explanation remains faithful to the fundamental ideas of the concept being described; and (3) \textit{Effectiveness of Analogy or Metaphor}, which evaluates how well the chosen metaphor aids comprehension and aligns with the original concept.

\subsection{Reading Comprehension of Metaphors (RCM)}

RCM tasks assess a model’s ability to interpret metaphors in literary, journalistic, and conceptual texts. These tasks evaluate not only the recognition of metaphorical language but also the capacity to establish meaningful connections between source and target domains and explain how metaphors shape meaning and perception. The complexity of metaphor interpretation makes these tasks particularly useful for benchmarking conceptual understanding in large language models.

Each task presents a metaphorical passage, requiring the model to detect the metaphor, analyze the underlying conceptual mappings, and articulate its interpretative significance. For example, in the passage \emph{``His words cut deep, leaving wounds that time struggled to heal,''} the model must recognize how the metaphor of physical injury conveys emotional pain and lasting psychological effects. Similarly, in \emph{``The economy is a house of cards, ready to collapse at the slightest disturbance,''} the model must infer how the imagery of structural fragility conveys economic instability. 

Performance in RCM tasks is evaluated based on three key criteria. The first is \textit{precision in metaphor identification}, which measures the model’s ability to accurately detect metaphorical expressions and distinguish them from literal statements. The second is \textit{completeness of source-target mapping}, assessing whether the model correctly links metaphorical source domains, such as fire, seeds, or storms, to their respective target domains, such as motivation, knowledge, or political unrest. Finally, \textit{depth of interpretive insight} is used to evaluate how effectively the model explains the metaphor’s impact on meaning and perception, considering logical coherence and conceptual depth.

\section{Experiments}
\label{sec:exp}

We evaluate the CMT-based prompting approach against baseline prompts across the four introduced distinct categories of tasks.

\subsection{Evaluation}
\label{sec:evaluation}

Each task was evaluated based on three predefined criteria, with scores ranging from 1 to 5. Independent evaluations by multiple annotators ensured consistency, with inter-rater agreement assessed for reliability.

To systematically compare responses, we employed {Llama3.3} (70B) as an expert evaluator. This model assessed both baseline and CMT responses against predefined criteria, including accuracy, coherence, and relevance.

The evaluator received detailed instructions, including the task description, baseline and CMT responses, and scoring criteria. It assigned scores from 1 to 5 for each criterion, provided justifications, and identified the superior response or noted ties. The evaluation prompt is shown in Figure~\ref{fig:eval}.

This structured approach ensures fairness and consistency while leveraging {Llama3.3}'s capabilities for insightful assessments of CMT’s impact.

\begin{figure}[htb]
\begin{lstlisting}
You are an expert evaluator. Your task is to assess two responses (Baseline vs. CMT) based on specific evaluation criteria. Provide scores and a rationale for each criterion.

Task Description:
{[Task description provided in the JSON]}

Baseline Model Response:
{[Baseline response from the native model]}

CMT-prompted Model Response:
{[Response from the CMT-prompted model]}

Evaluation Criteria:
- {[Criterion 1]}
- {[Criterion 2]}
- {[Criterion 3]}
...

Instructions:
1. For each criterion, assign a score from 1 (poor) to 5 (excellent).
2. Provide a short rationale for each score to justify your evaluation.
3. Conclude which response is better overall (Baseline or CMT) or indicate a tie.
\end{lstlisting}
\caption{Prompt for evaluation with Llama3.3\label{fig:eval}}
\end{figure}

\subsection{Experimental Setup}

We selected representative tasks from each category. These tasks were tested across four large language models: {Llama3.2} with 3B parameters, {Phi3} with 3.8B parameters, {Gemma2} with 2B parameters, and  {Mistral} with 7B parameters. Each model was tested in both its baseline configuration and with CMT-enhanced prompting. The responses were evaluated based on the predefined criteria, and results were averaged across all tasks within each category.

\subsection{Experimental Results}

\begin{figure*}[h]
    \centering
    \includegraphics[width=0.245\textwidth]{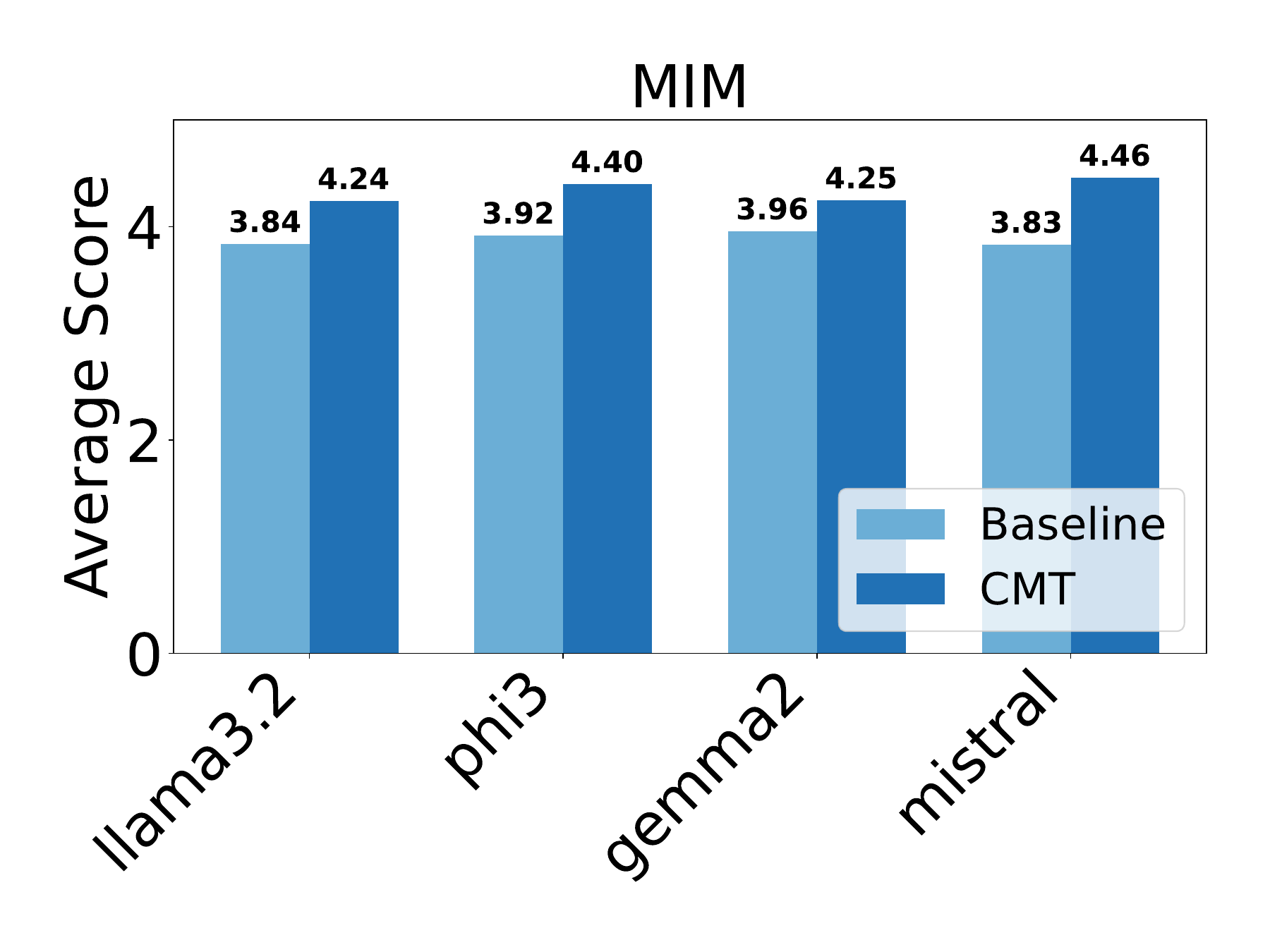}
    \includegraphics[width=0.245\textwidth]{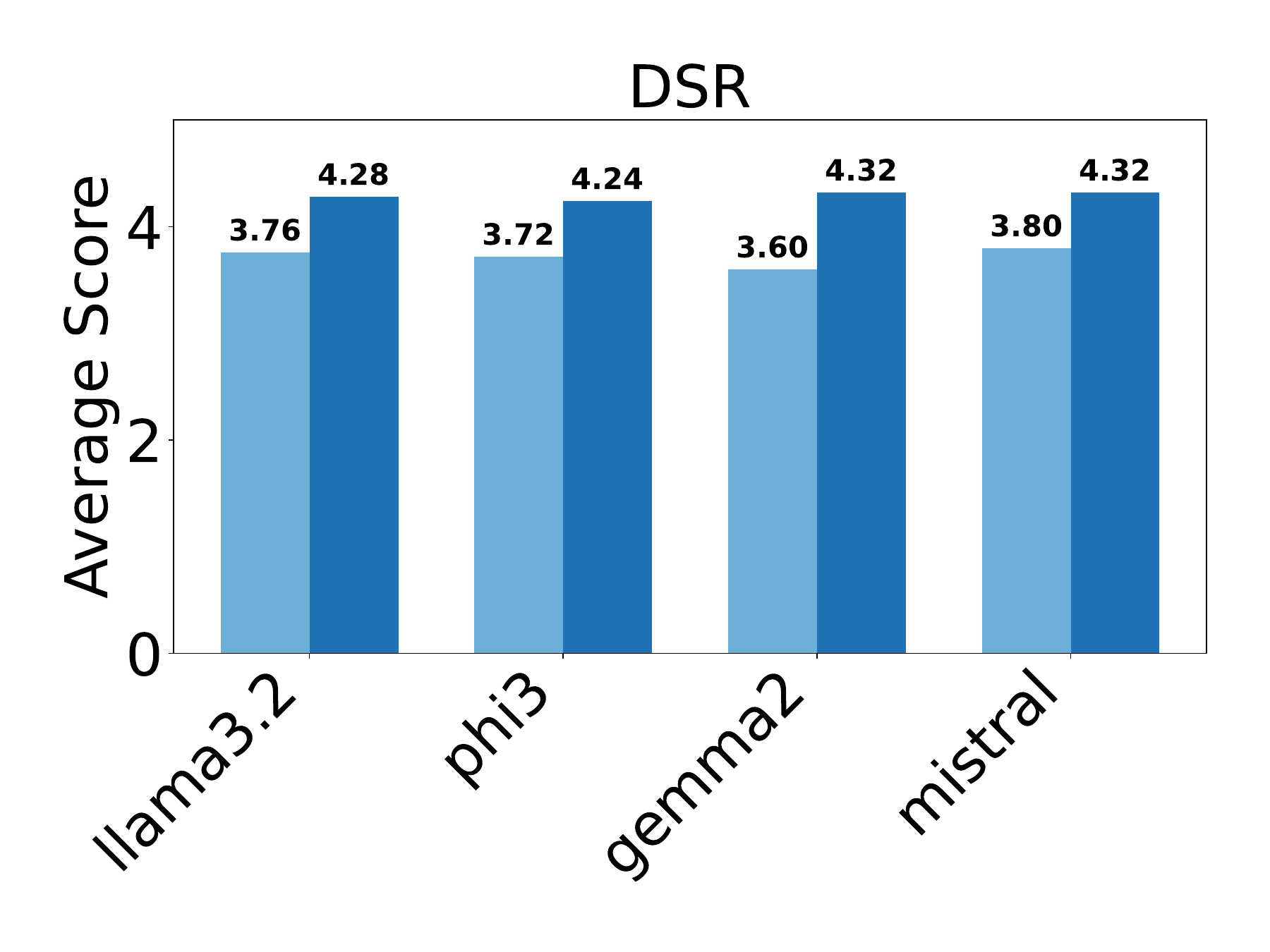}
    \includegraphics[width=0.245\textwidth]{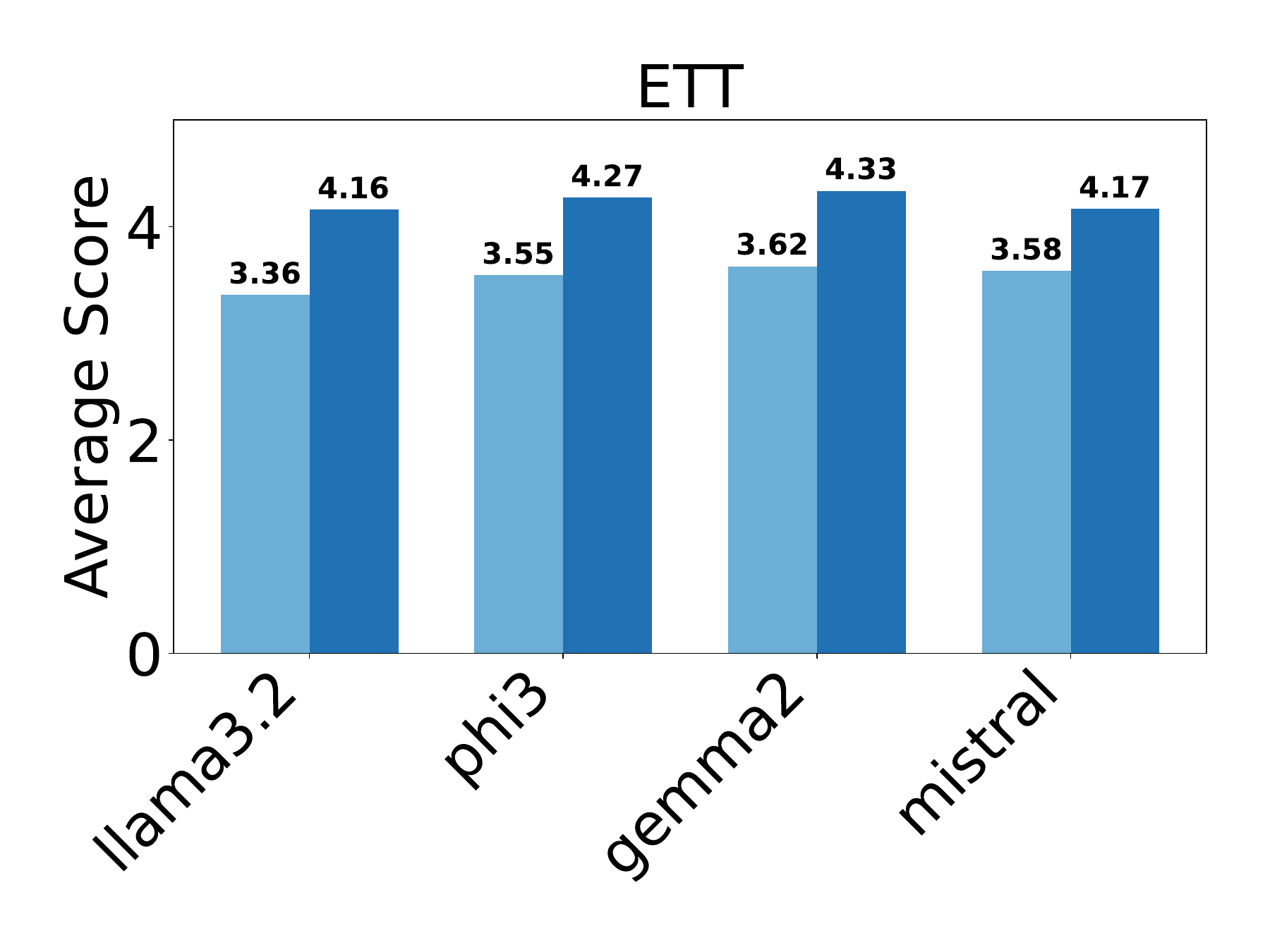}
    \includegraphics[width=0.245\textwidth]{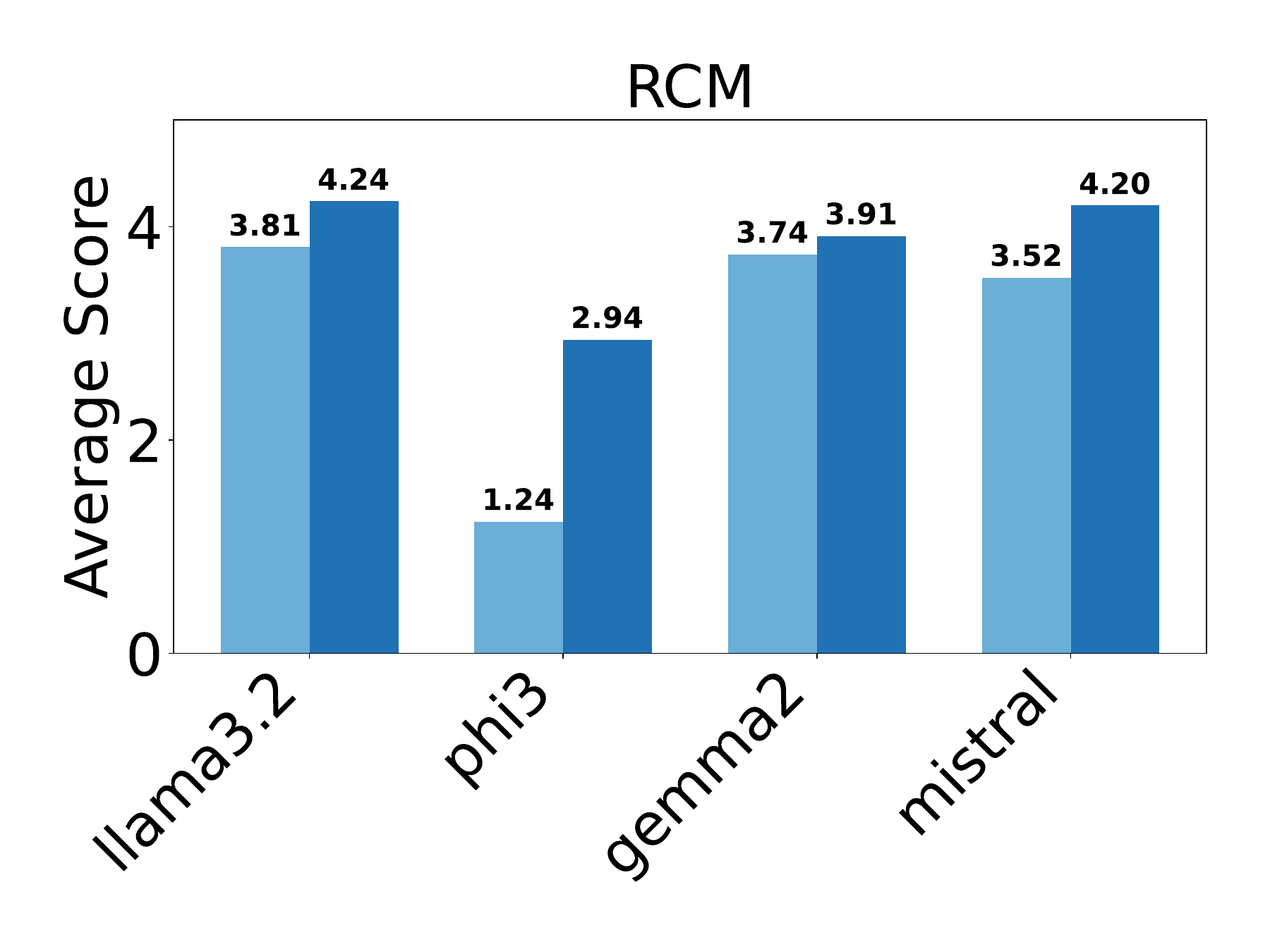}
\caption{Comparison of baseline and CMT-enhanced LLM performance across task categories. The four categories—MIM, DSR, ETT, and RCM—are shown from left to right, with average scores per task class.}
    \label{fig:results}
\end{figure*}

Figure~\ref{fig:results} presents the average baseline- and CMT-LLM scores for the four LLMs in MIM, DSR, ETT, and RCM.  

The results of MIM, see Figure~\ref{fig:results} (left), show that CMT-enhanced models consistently outperform baselines in metaphor interpretation. CMT prompting is particularly effective for complex conceptual mappings, such as economic and physical burden metaphors, enabling more precise source-target alignment. For instance, in "The economy is entering a deep freeze," CMT improved recognition of layered metaphorical structures. While baselines handled conventional expressions well, CMT enhanced explanatory depth, especially for nuanced metaphors. Models like Mistral and Phi3 showed the greatest improvements in coherence and interpretative insight. Metaphors involving movement, transformation, and containment saw the highest gains from CMT's structured reasoning. Even where baselines performed well, CMT often provided additional conceptual clarity without adding unnecessary complexity. Minor anomalies, such as unusually high CMT-enhanced scores, suggest potential refinements to evaluation methods.

The evaluation of DSR, see Figure~\ref{fig:results} (2nd from left), shows that CMT-prompted models generally match or surpass baselines, with notable improvements in specific cases. CMT enhances clarity and structured reasoning, particularly in tasks requiring abstract conceptual mapping. For explanations of the immune system and blockchain operations, CMT provided richer metaphorical insights, aiding comprehension. {Phi3}, {Gemma2}, and {Mistral} frequently outperformed their baselines, especially in analogy-driven tasks. Financial markets, neural adaptation, and encryption showed significant gains, where metaphorical reasoning improved conceptual understanding. While {Llama3.2} exhibited some fluctuations, most models benefited from CMT in structuring domain-specific explanations. CMT's ability to foster intuitive reasoning is particularly valuable in technical subjects that rely on conceptual transfer. Mixed results in economic inflation and competition dynamics suggest room for refining metaphor application.

The evaluation of ETT, see Figure~\ref{fig:results} (2nd from right), highlights CMT's effectiveness in improving clarity and comprehension. CMT-enhanced models consistently outperformed or matched baselines, particularly in tasks requiring conceptual mapping and intuitive explanations. Significant gains were observed in explaining neural networks, historical alliances, and artificial intelligence, where metaphorical reasoning provided an accessible framework. {Llama3.2} and {Phi3} showed notable improvements under CMT, demonstrating its value in structuring explanations. Metaphor-enhanced prompting simplified complex topics like economic inflation and stock markets, enhancing accessibility. While some tasks saw similar performance between baseline and CMT models, CMT generally facilitated more coherent and engaging explanations. Physics and biology particularly benefited from structured analogies, aiding conceptual transfer. Relating abstract principles to familiar experiences proved advantageous for teaching-oriented outputs. These findings reinforce CMT as a powerful framework for enhancing language model explanations.

The RCM evaluation, see Figure~\ref{fig:results} (right), demonstrates CMT’s effectiveness in metaphor interpretation and figurative language analysis. CMT-enhanced models consistently outperformed or matched baselines across various metaphorical contexts. {Llama3.2}, {Gemma2}, and {Mistral} showed substantial gains in understanding complex metaphors when guided by CMT-based reasoning. Shakespeare’s “All the world’s a stage” and similar constructs particularly benefited from CMT’s structured approach to meaning extraction. CMT-enhanced models exhibited deeper comprehension of connotations and abstract associations, improving metaphor interpretation. In political and economic metaphors, CMT models provided more nuanced and contextually appropriate readings. {Phi3} demonstrated strong improvements, highlighting CMT’s adaptability to different linguistic frameworks. While some results showed parity between baseline and CMT models, CMT generally facilitated stronger metaphor deconstruction and contextual reasoning. Structured conceptual mappings enriched interpretation, yielding more insightful analyses.

\section{Conclusion}
\label{sec:cons}

Among individual models, Llama3.2 benefits the most from CMT, showing strong improvements across ETT, DSR, and parts of RCM, reinforcing its ability to generate structured responses. Phi3 displays significant gains in ETT and moderate improvements in RCM, suggesting CMT enhances conceptual mapping in this model. Gemma2 performs well in structured reasoning but faces challenges in metaphor-related tasks, particularly in RCM, where its scores remain inconsistent. Mistral demonstrates the most unpredictable performance—while it benefits in DSR and some ETT cases, its RCM performance remains inconsistent, and MIM shows limited gains. These findings indicate that CMT is particularly impactful for tasks requiring structured explanations, conceptual abstraction, and cognitive organization, reinforcing its potential as a powerful prompting strategy for enhancing LLM capabilities in reasoning and teaching-based tasks. Furthermore, the varying degrees of improvement across models suggest that CMT’s effectiveness depends not only on the nature of the task but also on the underlying architecture and training dynamics of each LLM.

\section{Outlook}
\label{sec:outlook}

Future work will explore extending the CMT framework to real-world, multimodal tasks and evaluating its applicability to emerging challenges in explainability, reasoning, and collaborative AI systems. 
Expanding the benchmark dataset to include more complex real-world tasks, such as multi-turn dialogue or cross-domain reasoning, could provide deeper insights into the scalability of CMT prompting. 
Finally, integrating user feedback into the prompt refinement process may offer valuable guidance on optimizing CMT strategies for both interpretative and technical tasks. By addressing these challenges, we aim to unlock the full potential of conceptual metaphor-driven prompting in the next generation of language models.

\balance
\bibliographystyle{unsrt}

\begin{thebibliography}{10}

\bibitem{Lakoff1}
George Lakoff and Mark Johnson.
\newblock {\em Metaphors We Live By}.
\newblock University of Chicago Press, Chicago, IL, 1980.

\bibitem{Lakoff2}
George Lakoff.
\newblock The contemporary theory of metaphor.
\newblock In {\em Metaphor and Thought}, pages 202--251. Cambridge University
  Press, Cambridge, 2 edition, 1993.

\bibitem{brown2020language}
Tom~B. Brown, Benjamin Mann, Nick Ryder, Melanie Subbiah, Jared Kaplan,
  Prafulla Dhariwal, Arvind Neelakantan, Pranav Shyam, Girish Sastry, Amanda
  Askell, Sandhini Agarwal, Ariel Herbert-Voss, Gretchen Krueger, Tom Henighan,
  Rewon Child, Aditya Ramesh, Daniel~M. Ziegler, Jeffrey Wu, Clemens Winter,
  Christopher Hesse, Mark Chen, Eric Sigler, Mateusz Litwin, Scott Gray,
  Benjamin Chess, Jack Clark, Christopher Berner, Sam McCandlish, Alec Radford,
  Ilya Sutskever, and Dario Amodei.
\newblock Language models are few-shot learners.
\newblock In {\em Neural Information Processing Systems (NeurIPS)}, volume~35,
  pages 24824--24837, 2022.

\bibitem{wei2022chain}
Jason Wei, Xuezhi Wang, Dale Schuurmans, Maarten Bosma, Brian Ichter, Fei Xia,
  Ed~H. Chi, Quoc~V. Le, and Denny Zhou.
\newblock Chain-of-thought prompting elicits reasoning in large language
  models.
\newblock In {\em Neural Information Processing Systems (NeurIPS) Workshop},
  volume~35, pages 24824--24837, 2022.

\bibitem{tree}
Shunyu Yao, Dian Yu, Jeffrey Zhao, Izhak Shafran, Tom Griffiths, Yuan Cao, and
  Karthik Narasimhan.
\newblock Tree of thoughts: Deliberate problem solving with large language
  models.
\newblock In {\em Neural Information Processing Systems (NeurIPS)}, volume~36,
  pages 11809--11822, 2023.

\bibitem{yao2022react}
Shunyu Yao, Jeffrey Zhao, Dian Yu, Nan Du, Izhak Shafran, Karthik~R.
  Narasimhan, and Yuan Cao.
\newblock React: Synergizing reasoning and acting in language models.
\newblock In {\em International Conference on Learning Representations (ICLR)},
  2023.

\bibitem{promptbreeder2022}
Chrisantha Fernando, Dylan Banarse, Henryk Michalewski, Simon Osindero, and Tim
  Rocktäschel.
\newblock Promptbreeder: Self-referential self-improvement via prompt
  evolution.
\newblock {\em Neural Information Processing Systems (NeurIPS) Workshop}, 2023.

\bibitem{ape}
Yongchao Zhou, Andrei~Ioan Muresanu, Ziwen Han, Keiran Paster, Silviu Pitis,
  Harris Chan, and Jimmy Ba.
\newblock Large language models are human-level prompt engineers.
\newblock In {\em International Conference on Learning Representations (ICLR)},
  2023.

\bibitem{opro}
Chengrun Yang, Xuezhi Wang, Yifeng Lu, Hanxiao Liu, Quoc~V Le, Denny Zhou, and
  Xinyun Chen.
\newblock Large language models as optimizers.
\newblock In {\em International Conference on Learning Representations (ICLR)},
  2024.

\bibitem{cp}
Oliver Kramer and Jill Baumann.
\newblock Unlocking structured thinking in language models with cognitive
  prompting.
\newblock In {\em European Symposium on Artificial Neural Networks - {ESANN}},
  pages 1--6, 2025.

\bibitem{prystawski2023psychologically}
Ben Prystawski, Paul Thibodeau, Christopher Potts, and Noah~D. Goodman.
\newblock Psychologically-informed chain-of-thought prompts for metaphor
  understanding in large language models.
\newblock In {\em Proceedings of the 45th Annual Conference of the Cognitive
  Science Society (CogSci 2023)}, 2023.

\bibitem{Comsa}
Iulia~M. Comsa, Julian~Martin Eisenschlos, and Srini Narayanan.
\newblock Miqa: {A} benchmark for inference on metaphorical questions.
\newblock In {\em Proceedings of the 2nd Conference of the Asia-Pacific Chapter
  of the Association for Computational Linguistics and the 12th International
  Joint Conference on Natural Language Processing, {AACL/IJCNLP} 2022 - Volume
  2: Short Papers, Online only, November 20-23, 2022}, pages 373--381.
  Association for Computational Linguistics, 2022.

\bibitem{Hicke}
Rebecca M.~M. Hicke and Ross~Deans Kristensen{-}McLachlan.
\newblock {SCIENCE} {IS} {EXPLORATION:} computational frontiers for conceptual
  metaphor theory.
\newblock In {\em Proceedings of the Computational Humanities Research
  Conference 2024, Aarhus, Denmark, December 4-6, 2024}, volume 3834 of {\em
  {CEUR} Workshop Proceedings}, pages 1105--1116. CEUR-WS.org, 2024.

\end{thebibliography}

\end{document}